\documentclass[a4paper, 10 pt, conference]{ieeeconf} 

\IEEEoverridecommandlockouts
\usepackage{amsmath}
\usepackage{amsfonts}
\usepackage{bm}
\usepackage{cleveref}
\usepackage{subcaption}
\usepackage{bbm}
\usepackage{siunitx}
\usepackage{vector}
\renewcommand{\vec}[1]{\vect{#1}}

\usepackage{dsfont}
\usepackage{amssymb}

\usepackage[style=ieee,mincitenames=1,isbn=false,maxbibnames=99]{biblatex}
\AtEveryBibitem{
\ifentrytype{inproceedings}{
 	\clearlist{address}
 	\clearlist{publisher}
 	\clearname{editor}
 	\clearlist{organization}
 	\clearfield{url}  
 	\clearfield{doi}  
 	\clearfield{pages}  
 	\clearlist{location}
 }{}
 }
\usepackage{tikz}
\usetikzlibrary{positioning, arrows, patterns} 
\usetikzlibrary{shapes.geometric, shapes.misc, shapes.arrows}

\usepackage{algorithm}
\usepackage{algorithmicx}
\usepackage[noend]{algpseudocode}

\usepackage{booktabs}
\usepackage{multirow}
   
\usepackage{makecell}
\usepackage{mathtools}

\usepackage{etoolbox}
\makeatletter
\patchcmd{\@makecaption}
  {\scshape}
  {}
  {}
  {}
\makeatother

\usepackage{array}

\newcommand{\T}{\textsc{T}}
\newcommand{\F}{\textsc{F}}
\newcommand{\A}{\textsc{A}}
\newcommand{\True}{\textsc{True}}
\newcommand{\False}{\textsc{False}}
\newcommand{\Arbitrary}{\textsc{Arbitrary}}

\usepackage{mathtools}

\usepackage[keeplastbox]{flushend}

\title{\LARGE \bf Interpretable Safety Validation for Autonomous Vehicles }

\author{Anthony Corso$^{1}$, and Mykel J. Kochenderfer$^{1}$%
\thanks{A. Corso and M. J. Kochenderfer are with the Aeronautics and Astronautics Department, Stanford University. e-mail: \{acorso,mykel\}@stanford.edu}%
}

\begin{document}
\maketitle
\thispagestyle{empty}
\pagestyle{empty}

\begin{abstract}
An open problem for autonomous driving is how to validate the safety of an autonomous vehicle in simulation. Automated testing procedures can find failures of an autonomous system but these failures may be difficult to interpret due to their high dimensionality and may be so unlikely as to not be important. This work describes an approach for finding interpretable failures of an autonomous system. The failures are described by signal temporal logic expressions that can be understood by a human, and are optimized to produce failures that have high likelihood. Our methodology is demonstrated for the safety validation of an autonomous vehicle in the context of an unprotected left turn and a crosswalk with a pedestrian. Compared to a baseline importance sampling approach, our methodology finds more failures with higher likelihood while retaining interpretability.

\end{abstract}
\vskip -0.2in

\section{Introduction}

A major challenge for autonomous driving is how to validate the safety of an autonomous vehicle (AV) before deploying it on the road. A common practice is to test the AV using an adversarial driving simulator that controls stochastic disturbances in the environment over time to find failures~\cite{koren2018adaptive, corso2019adaptive, abeysirigoonawardena2019generating}. The disturbance trajectories that lead to failure are often high-dimensional and can therefore be difficult to interpret. In this work, we try to address this problem by developing a technique for interpretable black-box safety validation. We seek to generate descriptions of disturbance trajectories that lead to failure so that the descriptions may be used to understand weaknesses in the autonomous system and produce many related failure examples for further analysis.

The safety validation problem we study involves finding failures of an autonomous system (system-under-test or SUT) that operates in a stochastic environment. The state of the SUT and the environment is $s \in \mathcal{S}$ and the disturbances $x \in \mathcal{X}$ are stochastic changes in the environment that can influence the SUT. For autonomous driving, the state is the pose of each agent and the internal state of the AV policy, while the disturbances may include the behavior of other drivers, the behavior of pedestrians, and sensor noise. A disturbances trajectory $\vec{x} = \{x_1, \ldots, x_{N}\}$ has a probability $p(\vec{x})$ which models the stochastic elements of the environment. The set of all trajectories that end in a failure of the AV is denoted $E$ and $\vec{x} \in E$ means that the sequence of disturbances $\vec{x}$ causes the SUT to fail.

Black-box falsification techniques try to find any $\vec{x}$ such that $\vec{x} \in E$. Falsification can be cast as an optimization problem over the space of disturbance trajectories~\cite{Donze2010robust} which can be solved using various optimization algorithms~\cite{Mathesen2019falsification,Akazaki2018falsification,zhao2003generating,zhang2018two,sankaranarayanan2012falsification}. These approaches do not scale well when $\vec{x}$ is very high-dimensional and failure examples might be extremely rare according to $p(\vec{x})$. Adaptive stress testing~\cite{lee2015adaptive,koren2018adaptive,corso2019adaptive,koren2019Efficient} tries to find the most likely failure by maximizing $p(\vec{x})$ subject to $\vec{x} \in E$ and frames the falsification problem as a Markov decision process so it can be solved using reinforcement learning. Failures can also be found using importance sampling~\cite{huang2019evaluation,okelly2018scalable,uesato2018rigorous,zhao2016accelerated}, or with rapidly exploring random trees~\cite{koschi2019computationally}. Unlike the previous approaches, or work seeks to find an interpretable description of the failure trajectory $\vec{x}$ rather than just a single example.

For an algorithm to be interpretable, its results must be readily understood by a human. Interpretable models must choose an output representation that is sufficiently expressive to compactly convey the desired information while having direct mappings to words and concepts. Decision trees use a branching set of rules to classify data \cite{breiman2017classification} or represent a reinforcement learning policy \cite{rodriguez2019interpretable}. Simple mathematical expressions can be used for regression~\cite{schielzeth2010simple}, governing equation discovery~\cite{brunton2016discovering} and as a reinforcement learning policy~\cite{hein2017particle}. Signal temporal logic (STL) is well-suited for high-dimensional time series data~\cite{lee2018interpretable,vazquez2017logical} because it describes logical relationships between temporal variables and has a natural-language description that is easily understood.

Our approach seeks to find a description $\varphi$ such that any disturbance trajectory that satisfies the description will cause the SUT to fail. Additionally, we want to search for the most likely failure example by maximizing the probability of the failure trajectories. These goals can be combined into the optimization problem 
\begin{equation}
\max_\varphi \quad \mathbb{E}_{\vec{x} \sim p(\vec{x} \mid \vec{x} \in \varphi)}[p(\vec{x})] \quad\textrm{s.t.}\quad \vec{x} \in E\\
\end{equation}
where $p(\vec{x} \mid \vec{x} \in \varphi)$ is the probability density of disturbance trajectories that satisfy the description. Taking inspiration from related work, we will represent $\varphi$ using STL and perform the optimization using genetic programming. 

We use our approach to find the most-likely failures of an autonomous vehicle in two driving scenarios: 1) An unprotected left turn scenario with a small discrete disturbance space, and 2) a crosswalk scenario with a continuous and high-dimensional disturbance space to show that the approach scales. The first scenario assumes that the disturbances are independent while the second scenario models the joint distribution over disturbance trajectories as a Gaussian process. In both scenarios, we find interpretable failure modes of the AV for several different initial conditions. Additionally, our approach finds an order of magnitude more failures than the importance sampling baseline and the failures that are found have a much higher likelihood. The main contributions of this paper are:
\begin{enumerate}
    \item An approach for generating interpretable trajectory descriptions.
    \item Application of the approach to finding the most likely failures of an autonomous vehicle. 
    \item A comparison to an existing validation approach for two driving scenarios.
\end{enumerate}

The paper is organized as follows. \Cref{sec:background} describes the necessary technical background for our approach. \Cref{sec:methods} gives details of the algorithm. \Cref{sec:experiments} describes two experiments with safety validation of autonomous vehicles. \Cref{sec:conclusion} concludes and gives future direction.

\section{Background}
\label{sec:background}
This section discusses signal temporal logic, context-free grammars, genetic programming for expression optimization, and Gaussian processes.

\subsection{Signal Temporal Logic}
\label{subsec:stl}
Signal temporal logic (STL) is a logical formalism that is used to describe the behavior of time-varying systems~\cite{maler2004monitoring,baier2008principles}. STL uses the basic logical propositions \emph{and} ($\land$), \emph{or} ($\lor$), and \emph{not} ($\neg$), as well as temporal propositions such as \emph{eventually} ($\lozenge$) and \emph{always} ($\square$). 
Temporal logic operators are evaluated on series of Booleans. Continuous and discrete time series data are converted to Boolean statements using the comparison operators $\leq$, $\geq$, and $=$. A time series $\vec{x}$ satisfies an STL expression $\psi$ if $\psi(\vec{x})$ evaluates to \True{}.

STL expressions can also include an explicit dependence on time. The proposition $\square_{[t_1, t_2]} (x < 10)$ means \emph{always $x < 10$ for $t\in [t_1, t_2]$} and $\lozenge_{[t_1, t_2]}(x = 13)$ means \emph{eventually $x = 13$ for $t\in [t_1, t_2]$}. The flexibility of STL and the ease with which STL expressions can be converted into natural language can make it a suitable choice for interpretable modeling of heterogeneous time series~\cite{lee2018interpretable,vazquez2017logical}.

\subsection{Context-Free Grammar}
\label{subsec:cfg}

A context-free grammar is a set of rules that define a formal language such as STL.  Each rule in the grammar is composed of an output type and a set of inputs consisting of types and symbols. To generate an expression, a starting type is chosen and types are recursively expanded until only symbols remain. 

The grammar for the STL language described in \cref{subsec:stl} is given in \cref{eq:stl_grammar} for a single time series variable $x$. The type $\mathbb{B}$ is used for Boolean scalars, $\mathbb{S}$ is for Boolean series, $\mathbb{T}$ is for time, and $\mathbb{X}$ is for values of $x$. The logical operators apply element-wise on series. The symbol $\mid$ separates rules on the same line and $:$ indicates a range of symbols. In this work, time is discrete and the variable $x$ may be discrete or continuous. When $x$ is continuous, $\mathbb{X}$ is sampled uniformly at random in the range $[x_{\min}, x_{\max}]$.
\begin{equation}
\label{eq:stl_grammar}
\begin{split}
    \mathbb{B} &\mapsto \mathbb{B} \land \mathbb{B} \mid \mathbb{B} \lor \mathbb{B} \mid \neg \mathbb{B} \mid \square_{[\mathbb{T},\mathbb{T}]}(\mathbb{S}) \mid \lozenge_{[\mathbb{T},\mathbb{T}]}(\mathbb{S}) \\
    \mathbb{T} &\mapsto 0:T_{\max} \\
    \mathbb{S} &\mapsto \mathbb{S} \land \mathbb{S} \mid \mathbb{S} \lor \mathbb{S} \mid \neg \mathbb{S} \mid x \leq \mathbb{X} \mid x \geq \mathbb{X} \mid x = \mathbb{X} \\
    \mathbb{X} &\mapsto [x_{\min},x_{\max}]
\end{split}
\end{equation}

\subsection{Genetic Programming}
\label{subsec:gp}
Genetic programming is a population-based optimization technique for trees such as expressions generated from a grammar~\cite{kochenderfer2019algorithms, koza1992genetic}. Trees are evaluated according to a fitness function to determine the quality of an individual. To start the optimization, a population of $N_{\rm pop}$ trees is randomly sampled. Then, the following operations are performed randomly at each iteration (or generation) until convergence:
\begin{itemize}
    \item \textbf{Reproduction:} The fittest tree is selected from a subset of the population and progresses to the next iteration.
    \item \textbf{Mutation:} A random node in the tree is replaced with a random tree of the same type from the grammar.
    \item \textbf{Crossover:} Two individuals are selected at random and mixed to create a child. A random subtree from the first individual replaces a random node of the same type in the second individual.
\end{itemize}

\subsection{Gaussian Processes}
\label{subsec:gp2}

A Gaussian process~\cite{williams2006gaussian} is a stochastic process where any finite set of sample points $\vec{t} = \left[ t_1, \ldots, t_m \right]$ have values $\vec{x} = \left[ x_1, \ldots, x_m \right]$ that are distributed according to a multivariate normal distribution $
    \vec{x} \sim \mathcal{N}\big( \vec{\mu}(\vec{t}), \vec{K}(\vec{t}, \vec{t}) \big)$
where $\mu_i(\vec{t}) = \mu(t_i)$ for mean function $\mu$ and $K_{ij}(\vec{t}, \vec{t}') = k(t_i, t'_j)$ for kernel function $k$. A common choice for $k$ is the squared exponential kernel with variance $\sigma^2$ and characteristic length $\ell$ given by $
    k(t_1, t_2) = \sigma^2 {\rm exp}\left( - (t_1 - t_2)^2 / 2 \ell^2\right)$.
One strength of Gaussian processes is the ability to easily compute a posterior distribution after observing some values $\vec{x}^o$ at observation points $\vec{t}^o$.

To apply linear constraints on a Gaussian process, we use the procedure of \citeauthor{jidling2017linearly}~\cite{jidling2017linearly} where they apply a transformation to sample points from a truncated multivariate normal distribution. Sampling from a truncated multivariate distribution can be done efficiently with the minimax tilting approach~\cite{botev2017normal}. For notational convenience, we write a Gaussian process with mean $\mu$, kernel $k$, observed points $\vec{x}^o$, lower bound linear constraints $\vec{l}$ and upper bound linear constraints $\vec{u}$ as $
    \mathcal{GP}\big(\mu, k, \vec{x}^o, \vec{l}, \vec{u}\Big)$.

\section{Methods}
\label{sec:methods}
This section provides details for our approach to interpretable validation. We overview the validation algorithm and then discuss how expressions and trajectories are sampled.

\subsection{Algorithm Overview}
\label{subsec:alg_overview}

The procedure for interpretable safety validation is summarized in \cref{alg:interpretable}. The algorithm takes as input a grammar $\mathcal{G}_{\rm STL}$ that specifies the STL rules (\cref{eq:stl_grammar}) and a cost function $c$ which can be any function that leads to failures when minimized. In this work, $c(\vec{x}) = -p(\vec{x}) \mathds{1}\{ \vec{x} \in E\}$ so that trajectories that end in failure and have high-likelihood correspond with low cost. A population of expressions $\varphi_j$ are sampled from the grammar (line \ref{line:if_sample_grammar}) and then the algorithm iterates until the computational budget is exhausted. On each iteration, the cost for each expression $\varphi_j$ is computed by evaluating $c(\vec{x})$ for $N$ trajectories that satisfy $\varphi_j$ and taking the average (line \ref{line:if_compute_avg}). The details of how trajectories are sampled are given in \cref{subsec:stoch_constr,subsec:mvts}. Then, we perform the operations of genetic programming: reproduction, crossover and mutation, to evolve the population of expressions to a lower cost (lines \ref{line:if_selection} to \ref{line:if_mutation}). After looping, return the highest performing description (line \ref{line:if_return}).

\begin{algorithm}
\caption{Interpretable Validation} \label{alg:interpretable}
\begin{algorithmic}[1]
    \Function{InterpretableValidation}{$\mathcal{G}_{\rm STL}$, $c$}
    \State Sample $\{ \varphi_1, \ldots, \varphi_M \}$ from $\mathcal{G}_{\rm STL}$ \label{line:if_sample_grammar}
    \Loop
        \For{each expression $\varphi_j$} 
            \State Sample $\{\vec{x}_1, \ldots, \vec{x}_N \}$ from $p(\vec{x})$ s.t. $\vec{x}_i \in \varphi_j$\label{line:if_sample_spec}
            \State $c_{\varphi_j} \gets \frac{1}{N} \sum_{i=1}^N  c(\vec{x}_i)$ \label{line:if_compute_avg}
        \EndFor
        \State Select expressions $\{ \varphi_1, \ldots, \varphi_M \}$ based on $c_{\varphi_j}$\label{line:if_selection}
        \State Crossover expressions
        \State Mutate expressions \label{line:if_mutation}
    \EndLoop
    \State \textbf{return} best $\varphi_j$ \label{line:if_return}
    \EndFunction
\end{algorithmic}
\end{algorithm}
\subsection{Sampling Expressions}
\label{subsec:sampling_expr}
Expressions need to be sampled from the STL grammar during the initialization of the genetic programming optimization scheme and during the mutation procedure. Expressions are constructed as a tree of nodes where each node is a rule from the grammar. Each node in the expression tree is recursively expanded by choosing a rule uniformly at random from the grammar until each leaf has a terminal rule. A maximum depth of 10 is imposed on all expressions to keep them from becoming too large. The grammar is defined and sampled from using ExprRules.jl. For a more detailed analysis on sampling expressions from a grammar see \citeauthor{kochenderfer2019algorithms}~\cite{kochenderfer2019algorithms}.

\subsection{Converting Expressions into Stochastic Constraints}
\label{subsec:stoch_constr}
To generate trajectories that satisfy an expression, one could sample $\vec{x} \sim p(\vec{x})$ until a sample satisfies the given expression. If the expression places very restrictive requirements on the trajectory, however, then a large number of samples would be required to find one that satisfies the expression. To mitigate this problem, we construct linear constraints that, when imposed on samples from $p(\vec{x})$, enforce the satisfaction of the STL expression. There are efficient algorithms for sampling from certain linearly-constrained distributions~\cite{jidling2017linearly}, which makes this approach tractable. 

There are many different constraints that could enforce a time series to satisfy an STL expression. We seek to find the set of constraints that impose the minimum restriction on the time series while enforcing STL satisfaction, and choose one of these constraints at random. The procedure for sampling the minimally-restrictive constraints is given in \cref{alg:sampling_constraints}. The algorithm takes as input an expression $ex$, an initially empty vector $V$ to store the linear constraints, and the desired output of the expression $out$. Depending on the type of the expression, $out$ can be a scalar or a vector whose values can be \True{}, \False{}, or \Arbitrary{}, where \Arbitrary{} means that there is no restriction placed on what the expression should evaluate to. If the expression is a leaf (e.g. a direct comparison such as $x < 3$), then the expression and the output are stored as a pair to the vector of constraints (line \ref{line:sc_push}). If the expression is not a leaf, then it can be decomposed into subexpressions (e.g. $x < 3 \ \lor \ x > 1$ yields two subexpressions $x < 3$ and $x > 1$) (line \ref{line:sc_subexpr}). The output of each of those subexpressions is determined by the output of their parent expression (line \ref{line:sc_subexpr_out}). A set of subexpression outputs is sampled from the rules shown in \cref{tab:inverse_propositions}.  For each subexpression, the algorithm recurses (line \ref{line:sc_recurse}) until all leaf nodes of the original expression are converted to constraints. The next section describes how to use those constraints to sample a disturbance trajectory from certain distributions.

\begin{algorithm}
\caption{Sampling constraints for STL satisfaction}
    \label{alg:sampling_constraints}
\begin{algorithmic}[1]
    \Function{SampleConstraints}{$ex$, $out$, $V$}
    \If{ $ex$ is a leaf}
        \State push($V$, ($ex$, $out$)) \label{line:sc_push}
    \Else
        \State $E \gets$ SubExpressions($ex$) \label{line:sc_subexpr}
        \State $O \gets$ SubExpressionOutputs($ex$, $out$) \label{line:sc_subexpr_out}
        \For{$i$ in 1:length($E$)}
            \State SampleConstraints($E[i]$, $O[i]$, $V$) \label{line:sc_recurse}    
        \EndFor
    \EndIf
    \EndFunction
\end{algorithmic}
\end{algorithm}

\begin{table}
    \centering
    \caption{The minimally-restrictive set of subexpression outputs for the desired output of a parent expression. \T, \F, and \A \ are \True{}, \False{}, and \Arbitrary{}. To place the minimal amount of restriction on the time series, an output of \Arbitrary{} is used whenever possible.}
    \label{tab:inverse_propositions}
    \begin{tabular}{lll} 
        \toprule
        \textbf{Expression} & \textbf{Output} & \textbf{Subexpression Output}  \\
        \midrule
        \multirow{2}{*}{$\land$} & \True{} & (\T, \T) \\ & \False{} & (\F, \A) or (\A, \F) \\
        \midrule
        \multirow{2}{*}{$\lor$} & \True{} & (\T, \A) or (\A, \T) \\ & \False{} & (\F, \F) \\
        \midrule
        \multirow{2}{*}{$\neg$} & \True{} & \F \\ & \False{}  &  \T   \\
        \midrule
        \multirow{2}{*}{$\square_{[t_1, t_2]}$} & \True{} & \begin{tabular}{l} {[} \A, \ $\ldots$, \ $\underbrace{\T, \ \ldots, \ \T}_{[t_1,t_2]}$, \ \A, \  $\ldots$  {]} \end{tabular} \\ & \False{}  &  \begin{tabular}{l} {[} \A, \ $\ldots$, \ $\underset{\substack{\uparrow\\\mathclap{t \in [t_1,t_2]}}}{\F}$, \ \A, \  $\ldots$ {]} \end{tabular}   \\
        \midrule
        \multirow{2}{*}{$\lozenge_{[t_1, t_2]}$} & \True{} & \begin{tabular}{l} {[} \A, \ $\ldots$, \ $\underset{\substack{\uparrow\\\mathclap{t_1}}}{\F}$, \ \ldots, \ $\underset{\substack{\uparrow\\\mathclap{t \in [t_1,t_2]}}}{\T}$, \ \A, \  $\ldots$ {]} \end{tabular} \\  & \False{}  &  \begin{tabular}{c} {[} \A, \ $\ldots$, \ $\underbrace{\F, \ \ldots, \ \F}_{[t_1,t_2]}$, \ \A, \  $\ldots$ {]} \end{tabular} \\
        \bottomrule
    \end{tabular}
\end{table}

\subsection{Sampling Multivariate Time Series with Constraints}
\label{subsec:mvts}

Let $\left[ \vec{x}^{(1)}, \ldots, \vec{x}^{(m)} \right]$ be a multivariate times series with $m$ samples of $\vec{x} \in \mathbb{R}^n$. Suppose we are given a set of constraint functions $\vec{l}, \vec{u} : \mathbb{R} \to \mathbb{R}^n$ such that $l_j(t) \leq x_j(t) \leq u_j(t)$ and $l_j(t) \leq u_j(t)$ for $j=1,\ldots, n$. The most general model of this times series is given by the joint probability distribution 
\begin{equation}
     p(\vec{x}^{(1)}, \ldots, \vec{x}^{(m)} ) \quad \text{s.t.} \quad \vec{l}(t_i) \leq \vec{x}^{(i)} \leq \vec{u}(t_i) 
\end{equation}
where $t_i$ corresponds to the time of the sample point $\vec{x}^{(i)}$. The model $p$ captures the correlations in the data over time and between variables.

If time series samples and components of $\vec{x}$ are independent, then the probability distribution decomposes to
\begin{equation}
    p(\vec{x}^{(1)}, \ldots, \vec{x}^{(m)}) = \prod_{i=1}^m p(\vec{x}^{(i)}) = \prod_{i=1}^m \prod_{j=1}^n p(x_j^{(i)}) \text{.}
\end{equation}
If the probability model is uniform, the $j$th component of the sample at $t_i$ is distributed as
\begin{equation}
    x_j^{(i)} \sim \mathcal{U}(l_j(t_i), u_j(t_i))
\end{equation}
If the probability model of the $j$th component is normal with mean $\tilde{\mu}_j$ and variance $\tilde{\sigma}_j^2$, then the sample at $t_i$ is distributed as a truncated normal distribution
\begin{equation}
    x_j^{(i)} \sim \mathcal{TN}_{l_j(t_i),u_j(t_i)}(\tilde{\mu}_j, \tilde{\sigma}_j^2)
\end{equation}

When the time series is jointly distributed as a Gaussian process with mean function $\mu_j$ and kernel $k_j$ for the $j$th component, then samples are distributed according to
\begin{equation}
    \left[ x_j^{(1)}, \ldots, x_j^{(m)} \right] \sim \mathcal{GP}\Big(\mu_j, k_j, l_j(\vec{t}^o), l_j(\vec{t}^c), u_j(\vec{t}^c) \Big)
\end{equation}
The inequality constraints are separated into two sets: $\vec{t}^o$ where $l_j(\vec{t}^o) = u_j(\vec{t}^o)$ and $\vec{t}^c$ where  $l_j(\vec{t}^c) < u_j(\vec{t}^c)$.

\section{Experiments}
\label{sec:experiments}

This section describes experiments where interpretable safety validation is used to find failures of an autonomous vehicle in two driving scenarios: 1) an unprotected left turn, and 2) a vehicle approaching a crosswalk with a pedestrian. The system-under-test is an AV (referred to as the ego vehicle) that follows a modified version of the intelligent driver model (IDM)~\cite{Treiber2000congested}, a vehicle-following algorithm that tries to drive at a specified velocity while avoiding collisions with leading vehicles or pedestrians. The IDM is parameterized by a desired velocity of \SI{29}{m/s}, a minimum spacing of \SI{5}{m}, a maximum acceleration of \SI{3}{m/s^2} and a comfortable braking deceleration of \SI{2}{m/s^2}. In the left-turn scenario, the IDM is augmented with a rules-based algorithm for navigating the intersection that predicts the behavior of oncoming traffic based on turn signals and right-of-way norms.

In both driving scenarios, we aim to find the most likely failure (collision between the ego vehicle and another agent) of the SUT based on the models of the environment. The performance of our algorithm is evaluated on three metrics: 1) The rate of failures generated (evaluated from 500 trials), 2) the likelihood of the failure trajectories (evaluated from 500 failure trajectories), and 3) the interpretability of the STL expression. As a baseline, we use importance sampling to make finding failures more likely. The proposal distribution is chosen for each scenario separately.

In all of the experiments, the STL optimization was performed with genetic programming with a population of \num{1000} individuals over \num{30} generations, implemented in ExprOptimization.jl. The probability of reproduction was \num{0.3}, the probability of crossover was \num{0.3} and the probability of mutation was \num{0.4}.

\subsection{Scenario 1: Unprotected Left Turn}
\label{subsec:two_vehicle_safety}

\begin{figure}
    \centering
   \begin{subfigure}[t]{0.45\columnwidth}
        \centering
        \includegraphics[width=0.9\textwidth, trim={10cm 21.3cm 22cm 0},clip]{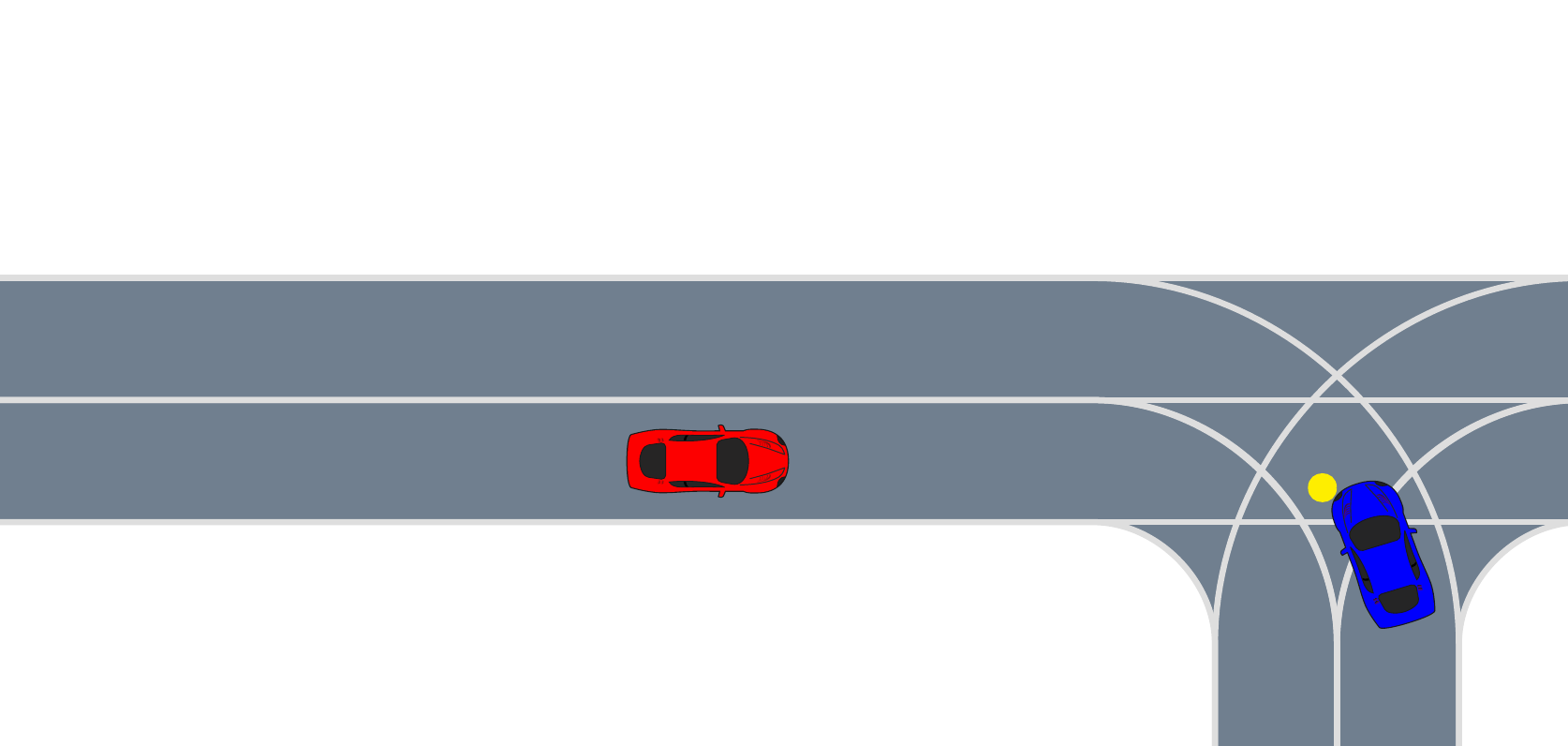}
    \end{subfigure}%
    \begin{subfigure}[t]{0.45\columnwidth}
        \centering
        \includegraphics[width=0.9\textwidth, trim={10cm 21.3cm 22cm 0},clip]{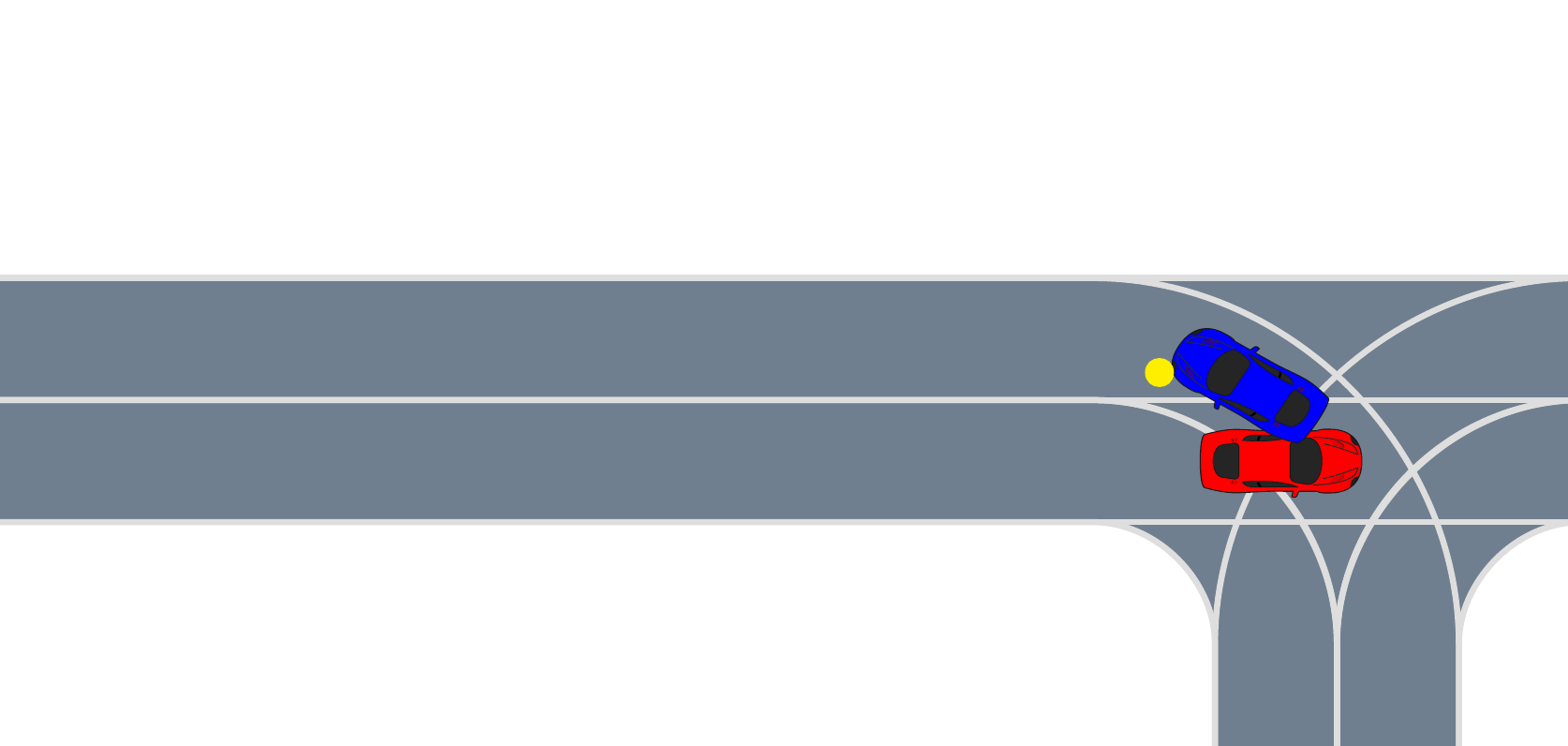}
    \end{subfigure}
    \caption{Collision for LT1 at $t=(\SI{1.08}{s}, \SI{1.98}{s})$.}
    \label{fig:2car_LT1}
    \vskip -0.2in
\end{figure}

\begin{figure}
    \centering
   \begin{subfigure}[t]{0.45\columnwidth}
        \centering
        \includegraphics[width=0.9\textwidth, trim={10cm 19.2cm 22cm 0},clip]{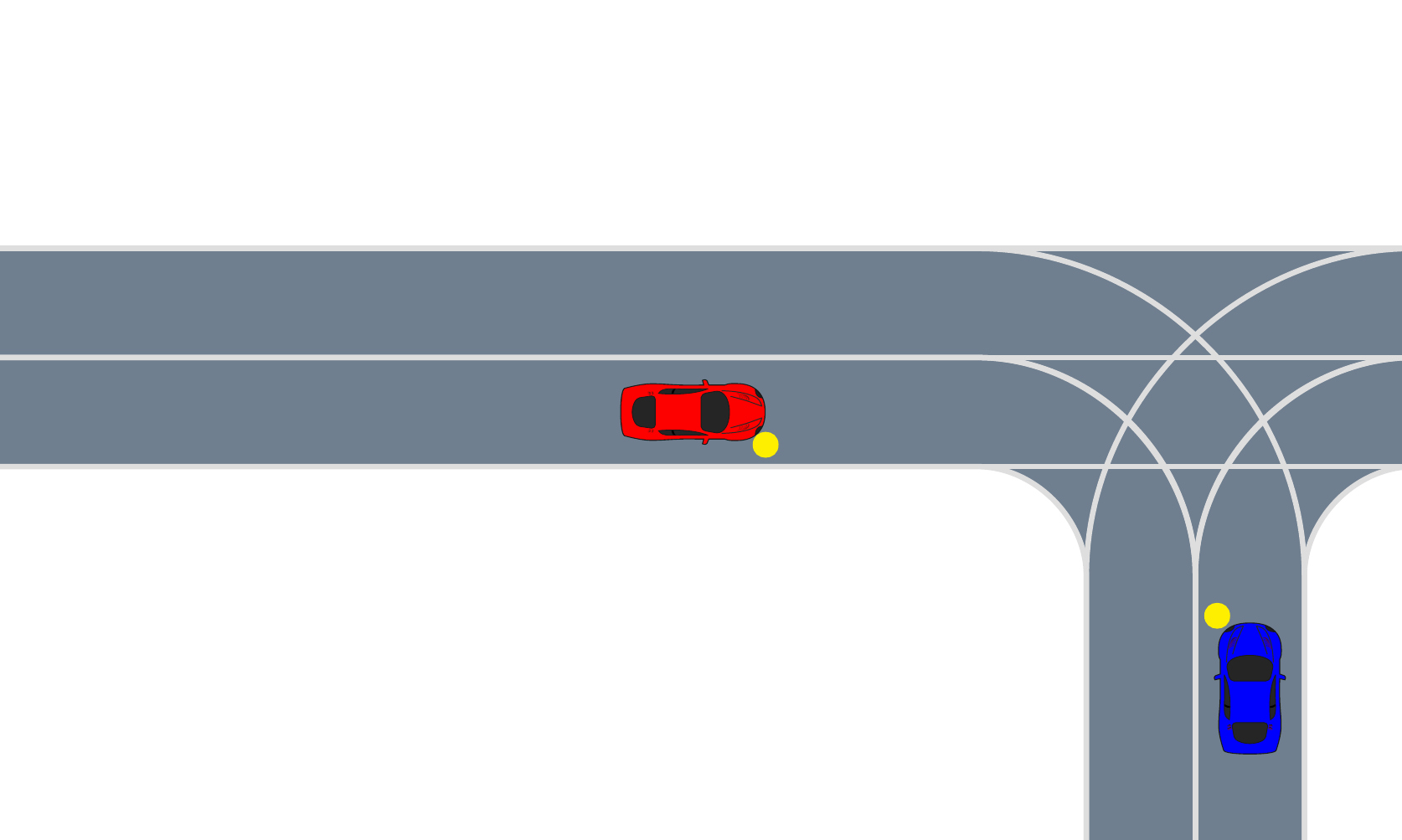}
    \end{subfigure}%
    \begin{subfigure}[t]{0.45\columnwidth}
        \centering
        \includegraphics[width=0.9\textwidth, trim={10cm 19.2cm 22cm 0},clip]{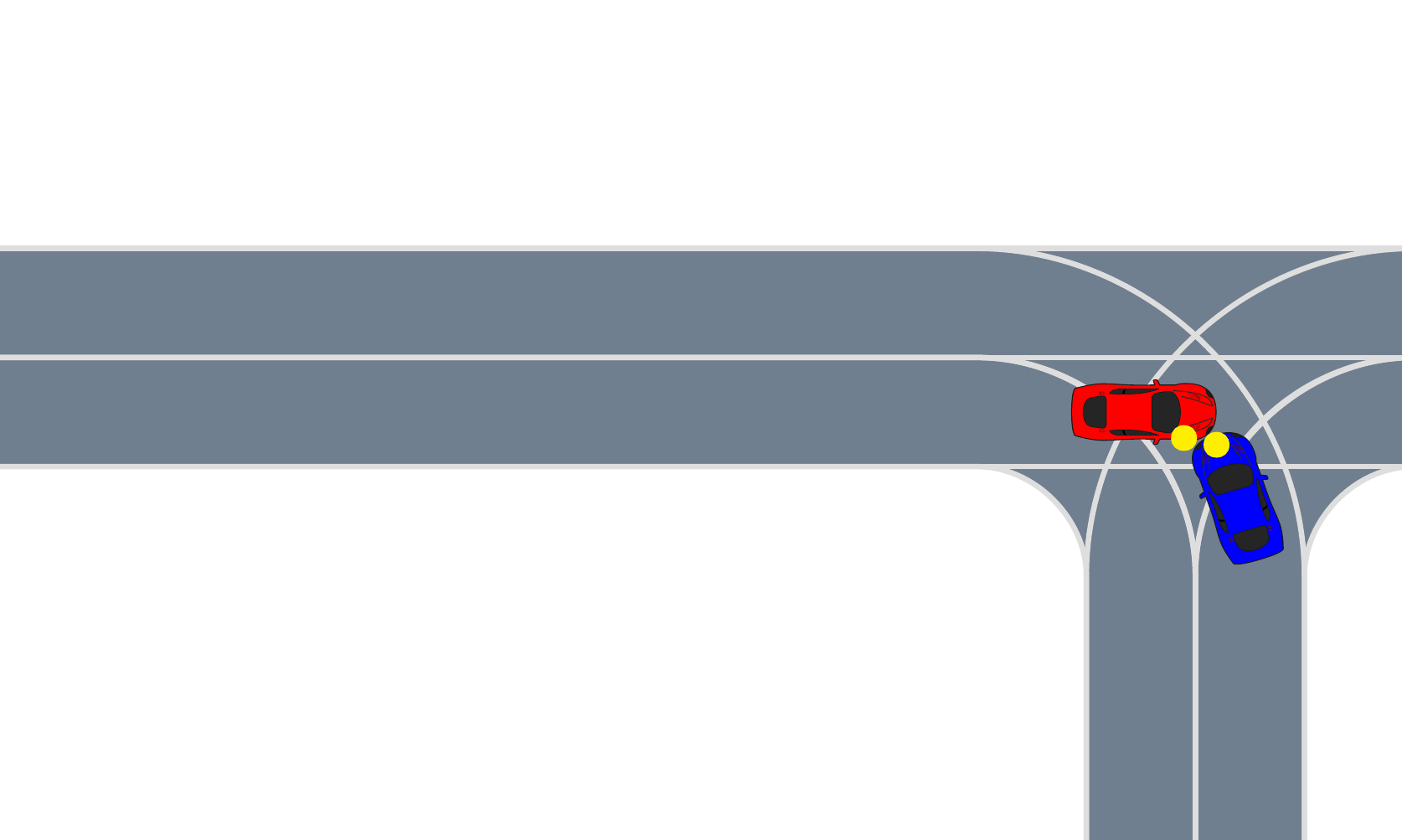}
    \end{subfigure}
    \caption{Collision for LT2 at $t=(\SI{0.72}{s}, \SI{1.26}{s})$.}
    \label{fig:2car_LT2}
    \vskip -0.2in
\end{figure}

\begin{figure}
\vskip -0.2in
    \centering
   \begin{subfigure}[t]{0.45\columnwidth}
        \centering
        \includegraphics[width=0.9\textwidth, trim={10cm 18.5cm 22cm 0},clip]{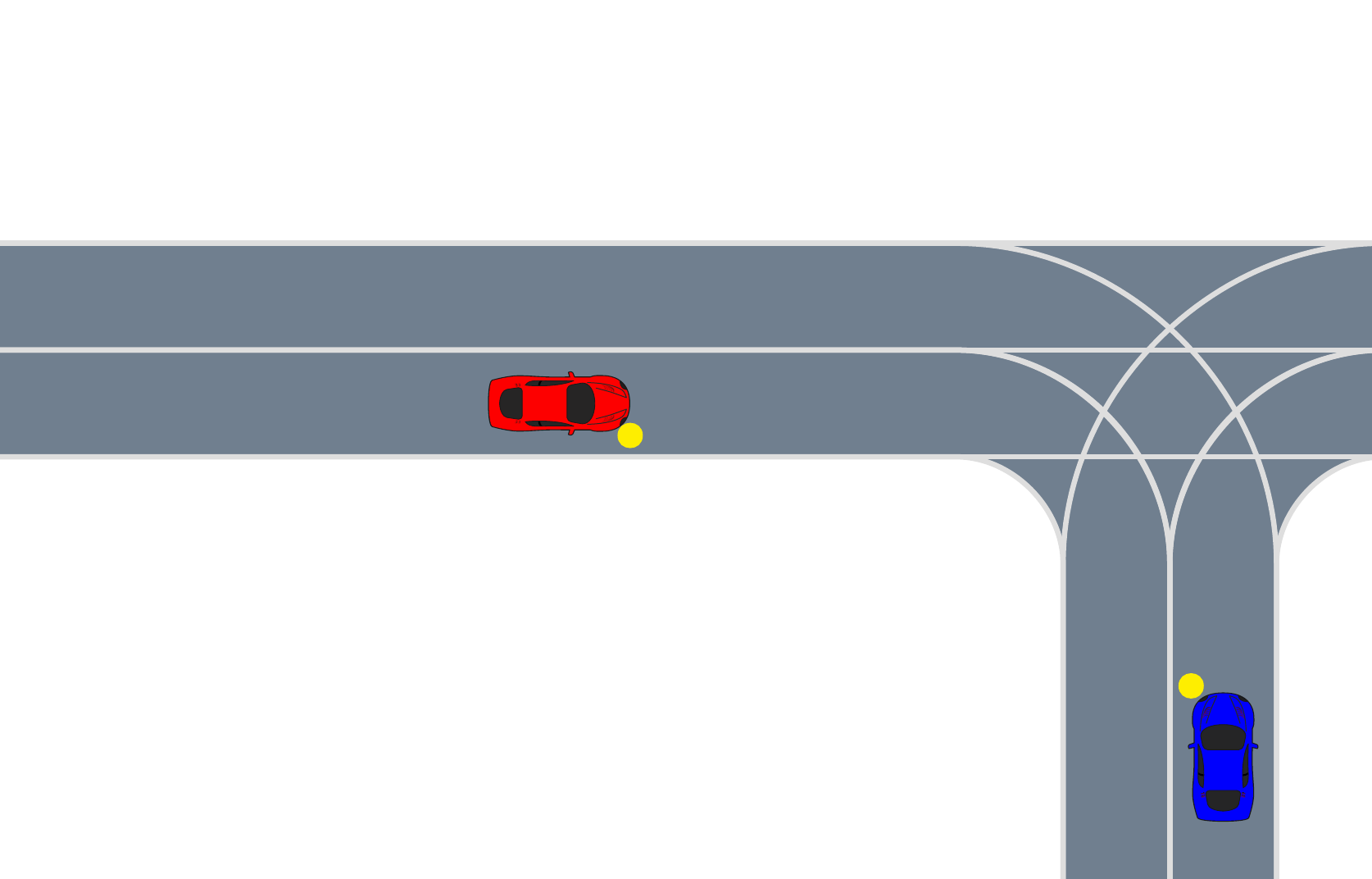}
    \end{subfigure}%
    \begin{subfigure}[t]{0.45\columnwidth}
        \centering
        \includegraphics[width=0.9\textwidth, trim={10cm 18.5cm 22cm 0},clip]{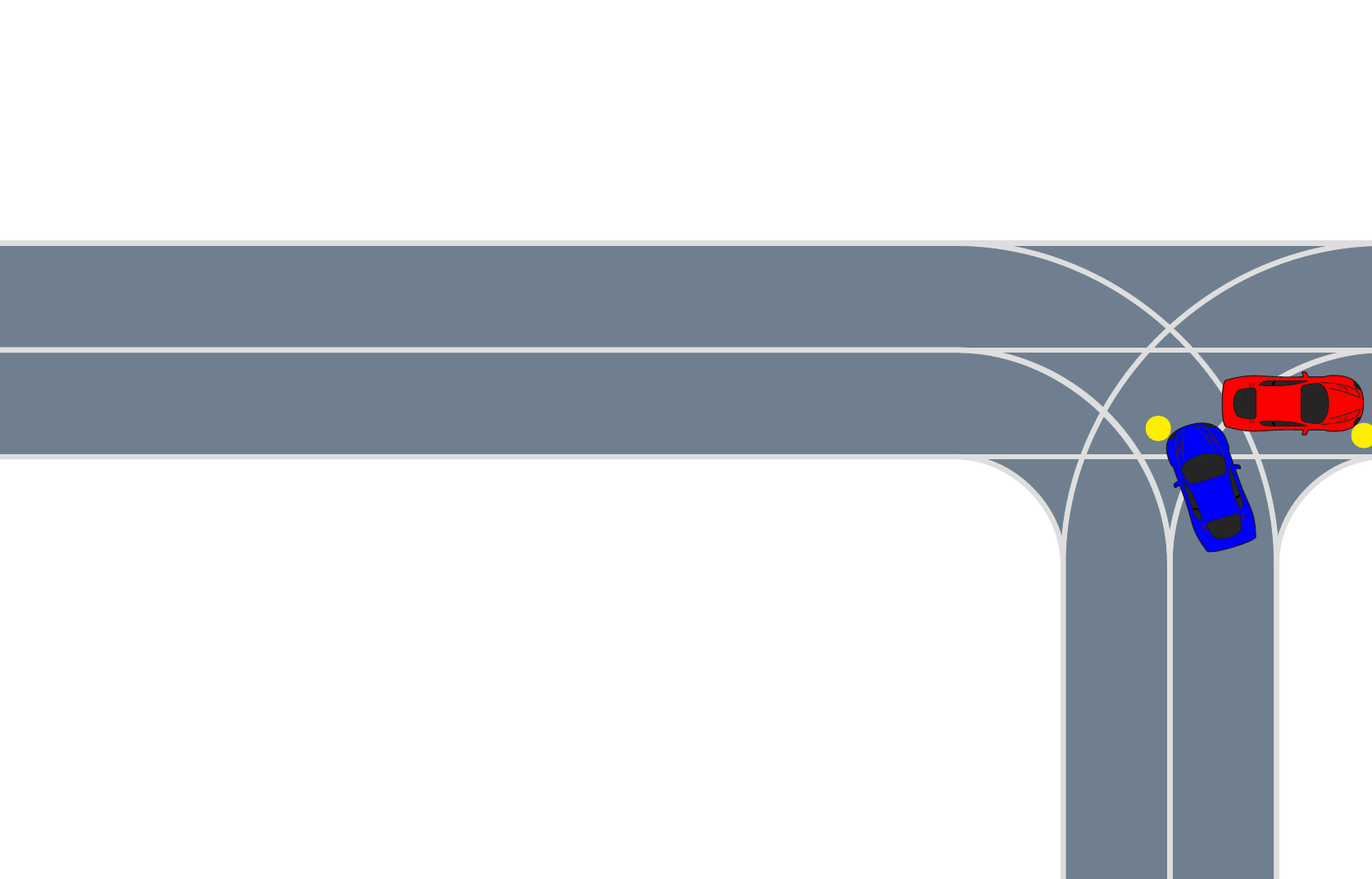}
    \end{subfigure}
    \caption{Collision for LT3 at $t=(\SI{0.90}{s}, \SI{1.62}{s})$.}
    \label{fig:2car_LT3}
\end{figure}

As shown in \cref{fig:2car_LT1,fig:2car_LT2,fig:2car_LT3}, the first scenario is an interaction between the ego vehicle (in blue), which is attempting to turn left, and one adversarial vehicle (in red) which is initially driving straight through the intersection. The yellow dot represents a turn signal. The adversarial vehicle has a discretized disturbance space with seven possible disturbances: no disturbance $\varnothing$, a medium slowdown $d_{\rm med}$, a major slowdown $d_{\rm maj}$, a medium speedup $a_{\rm med}$, a major speedup $a_{\rm maj}$, toggle turn signal $S$, and toggle turn intention $L$. The medium accelerations are $\pm \SI{1.5}{m/s^2}$ and the major accelerations are $\pm \SI{3.0}{m/s^2}$. The disturbances are added to the normal behavior of the vehicle dictated by the IDM, and have an associated probability with larger disturbances being less likely. The medium disturbances have probability \num{1e-2}, the major disturbances (including toggling turn signal and turn intention) have probability \num{1e-3}. The rest of the probability (\num{0.976}) is assigned to no disturbance. Under this probability model, failures are very unlikely to occur so we use importance sampling (IS) as a baseline. The proposal distribution was chosen to be uniform over the disturbances, a choice that significantly increases the likelihood of failure. The simulation timestep was set to $\Delta t = \SI{0.18}{s}$.

Three initial conditions of the left-turn (LT) scenario were investigated. Each initial condition is represented by the tuple ($s_{\rm ego}$, $v_{\rm ego}$, $s_{\rm adv}$, $v_{\rm adv}$) where $s$ is the distance from the center of the intersection and $v$ is the vehicle velocity.  The initial conditions and corresponding nominal behavior of the ego vehicle are given below.
\begin{itemize}
    \item LT1: (\SI{15}{m}, \SI{9}{m/s}, \SI{29}{m}, \SI{10}{m/s}). The nominal behavior is for the ego vehicle to take the left turn before the other vehicle arrives at the intersection.
    \item LT2: (\SI{15}{m}, \SI{9}{m/s}, \SI{29}{m}, \SI{20}{m/s}). The nominal behavior is for the ego vehicle to wait for other vehicle to pass through the intersection before turning left.
    \item LT3: (\SI{19}{m}, \SI{9}{m/s}, \SI{43}{m}, \SI{29}{m/s}). The nominal behavior is the same as in LT2.
\end{itemize}

The results for the three initial conditions are shown in \cref{tab:2car_results}. Trajectories that were sampled from the optimal STL expression produced at least an order of magnitude more failures than the rollouts from the IS distribution. This means that once the optimal expression is computed, it is very effective at generating failures. Additionally, we see that the average disturbance probability is much higher for the interpretable failures than for the IS failures. The interpretable failures place minimal constraints on the trajectories so many of the disturbances are selected according to the true model of the environment, leading to the relatively high likelihood of the trajectory. With importance sampling, the distribution causes rare disturbances to be chosen with regularity, even when they are not critical for causing a failure. Many rare events cause the IS trajectories to have very low likelihood.

Finally, we can interpret the generated failure expressions. For LT1, we see that the optimal STL expression enforces a major acceleration in the first two timesteps of the simulation. This acceleration allows the adversarial vehicle to just reach the ego vehicle as it completes its left turn as shown in \cref{fig:2car_LT1}. The ego vehicle initiated the turn based on the initial velocity of the other vehicle and did not predict such a significant speedup. For LT2, the generated expression has the adversarial vehicle toggle its turn signal but continue straight through the intersection as shown in \cref{fig:2car_LT2}. The ego vehicle expects the adversarial vehicle to turn, so it initiates the left turn and causes a collision. Lastly, for LT3, the generated expression has the adversary either turn on its turn signal or perform a major deceleration in the first two timesteps of the simulation (\cref{fig:2car_LT3}) leading to a similar failure as in LT2. From these experiments, we conclude that our approach is able to find interpretable expressions that reliable produce likely failures of the SUT.

\begin{table}
    \centering
    \caption{Comparing interpretable validation (IV) to importance sampling (IS) for three different initial conditions of the left-turn scenario.}
    \label{tab:2car_results}
    \begin{tabular}{@{}llll@{}} 
        \toprule
        \textbf{Method} & \textbf{Fail Rate} & \textbf{Likelihood} & \textbf{Expression}\\
        \midrule
        IV (LT1) & $0.968 \pm 0.003$ & $0.78 \pm 0.049$ & $\square_{[0, .36]} a_{\rm maj}$ \\
        IS (LT1) & $0.008 \pm 0.008$ & $0.11 \pm 0.099$ & - \\
        \midrule
        IV (LT2) & $0.994 \pm .001$ & $0.82 \pm 0.049$ & $\square_{[0, 0]} B$ \\
        IS (LT2) & $9.20 \pm 3.63$ & $0.15 \pm 0.13$ & - \\
        \midrule
        IV (LT3) & $0.164 \pm 0.004$ & $0.75 \pm 0.029$ & $\square_{[0, .36]} B \lor d_{\rm maj}$ \\
        IS (LT3) & $0.0018 \pm 0.0005$ & $0.15 \pm 0.13$ & - \\
        \bottomrule
    \end{tabular}
    \vskip -0.2in
\end{table}

\subsection{Scenario 2: Pedestrian Crossing}
\label{subsec:ped_crosswalk}
\begin{figure}
    \centering
  \begin{subfigure}[t]{0.5\columnwidth}
        \centering
        \includegraphics[width=0.9\columnwidth, trim={20cm 11cm 15cm 11cm},clip]{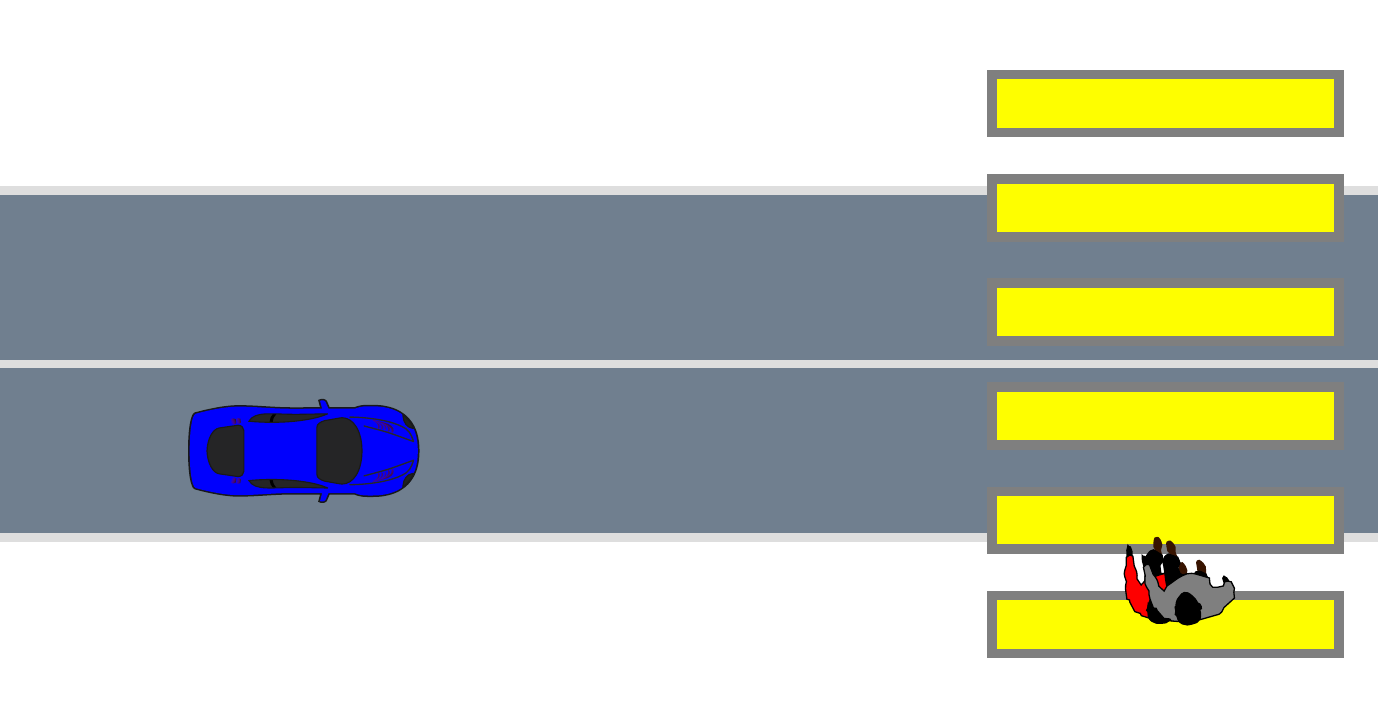}
    \end{subfigure}%
    \begin{subfigure}[t]{0.5\columnwidth}
        \centering
        \includegraphics[width=0.9\columnwidth, trim={20cm 11cm 15cm 11cm},clip]{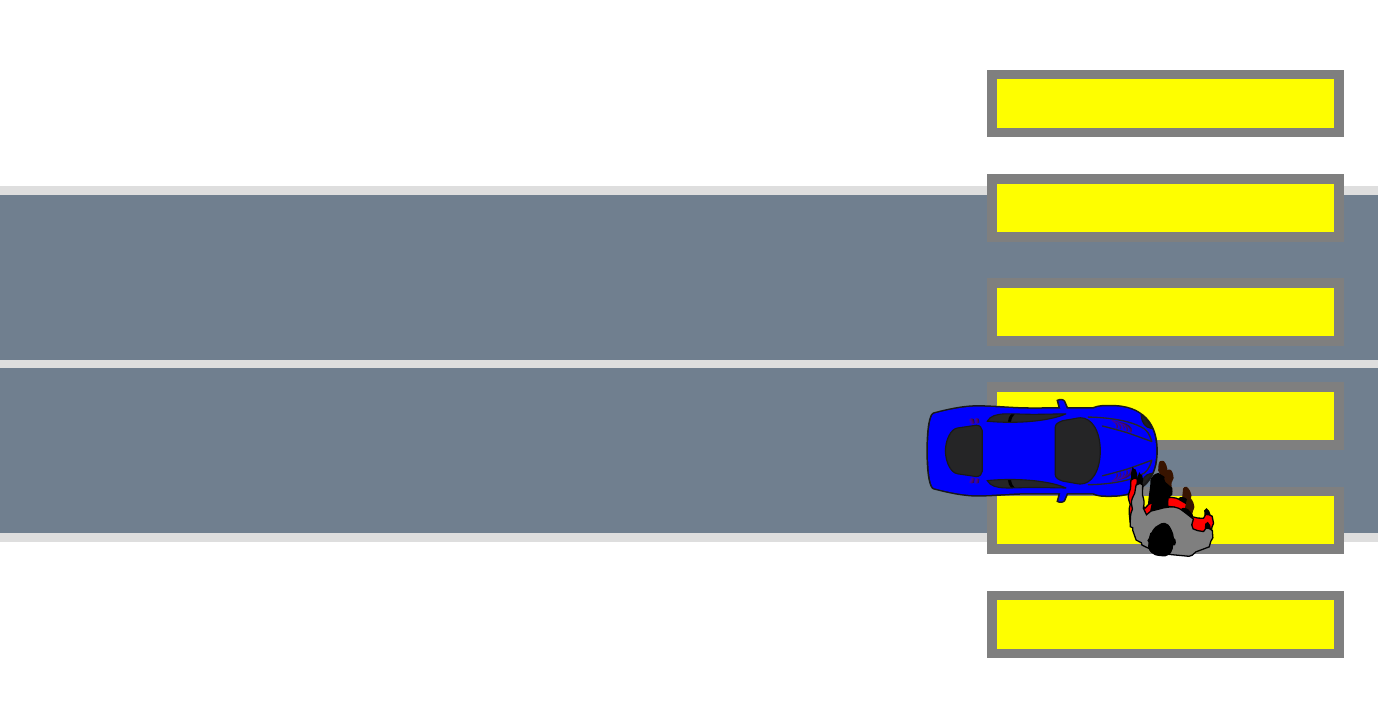}
    \end{subfigure}
    \caption{Collision for PC1 $t=(\SI{1.6}{s}, \SI{3.0}{s})$.}
    \label{fig:pedcar_PC1}
    \vskip -0.2in
\end{figure}

\begin{figure}
    \centering
  \begin{subfigure}[t]{0.5\columnwidth}
        \centering
        \includegraphics[width=0.9\columnwidth, trim={20cm 11cm 15cm 11cm},clip]{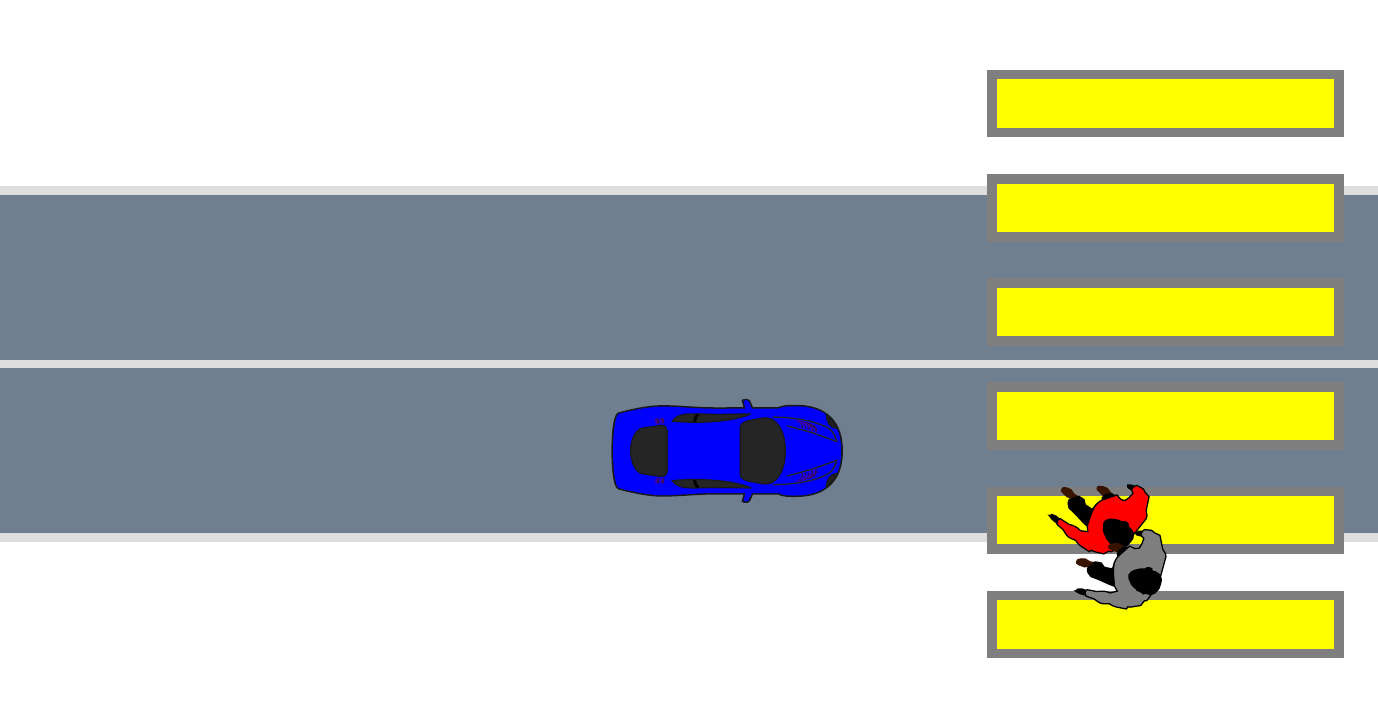}
    \end{subfigure}%
    \begin{subfigure}[t]{0.5\columnwidth}
        \centering
        \includegraphics[width=0.9\columnwidth, trim={20cm 11cm 15cm 11cm},clip]{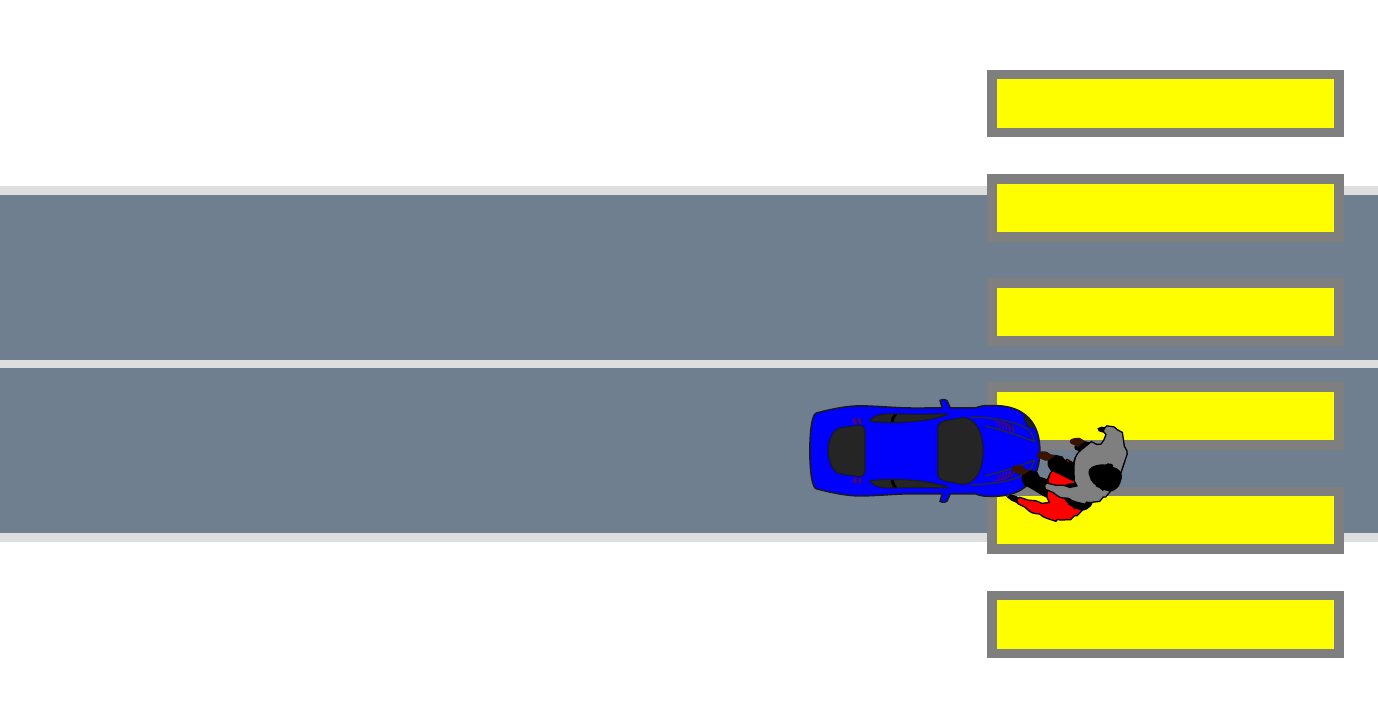}
    \end{subfigure}
    \caption{Collision for PC2 $t=(\SI{2.4}{s}, \SI{3.2}{s})$.}
    \label{fig:pedcar_PC2}
    \vskip -0.2in
\end{figure}

As shown in \cref{fig:pedcar_PC1,fig:pedcar_PC2}, the second driving scenario has an autonomous vehicle approach a crosswalk while a pedestrian tries to cross, as done by \citeauthor{koren2018adaptive}~\cite{koren2018adaptive}. The disturbances are the pedestrian acceleration $(a_x, a_y)$, the sensor noise on pedestrian position $(n_x, n_y)$, and the sensor noise on the pedestrian velocity $(n_{v_x}, n_{v_y})$. The sensor noise is modeled as independent samples from a zero-mean Gaussian with standard deviations $\sigma_{\rm pos}$ for the position noise in both directions and $\sigma_{\rm vel}$ for the velocity noise in both directions. The pedestrian acceleration in both directions was modeled as a zero-mean Gaussian process with a squared-exponential kernel. The characteristic length was chosen as $t=\SI{0.4}{s}$ so that the pedestrian cannot change acceleration too quickly. The standard deviation of the kernel is $\sigma_{\rm acc}$. The importance sampling distribution used the same disturbances models but doubled the standard deviation parameters to make larger disturbances more likely, and increase the failure rate). The comparisons in the STL expressions were sampled uniformly from [\SI{-2}{m/s^2}, \SI{2}{m/s^2}] for acceleration,  [\SI{-1}{m}, \SI{1}{m}] for position noise, and [\SI{-2}{m/s}, \SI{2}{m/s}] for velocity noise.

For these experiments, the vehicle starts \SI{35}{m} to the left of the crosswalk travelling at \SI{11.7}{m/s} and the pedestrian starts \SI{4}{m} below the center of the lane, moving at \SI{1.5}{m/s} to cross the street. The simulation timestep was set to $\Delta t = 0.2$. The correct behavior for the ego vehicle is to come to a safe stop before the crosswalk as the pedestrian crosses. The difference between pedestrian crosswalk (PC) scenarios is in the choice of probability parameters. The first set of parameters (PC1) makes pedestrian motion more volatile than sensor noise with $\sigma_{\rm acc} = \SI{1}{m/s^2}$, $\sigma_{\rm pos} = \SI{0.2}{m}$, $\sigma_{\rm vel} = \SI{0.5}{m/s^2}$. The second set (PC2) causes the noise to be a larger factor with $\sigma_{\rm acc} = \SI{1}{m/s^2}$, $\sigma_{\rm pos} = \SI{1}{m}$, $\sigma_{\rm vel} = \SI{1}{m/s^2}$.

The results of the experiments are shown in \cref{tab:pedcar_results}. In scenario PC1 and PC2, our approach found STL expressions that produced failures \num{100}\% of the time, an order of magnitude larger than the importance sampling approach. Additionally, the failures found via STL expressions had much higher likelihood than those found using importance sampling. In PC1 we see that the optimal STL expression prescribes a specific acceleration for the pedestrian that causes the pedestrian to enter the street and hit the car right as the car passes by the crosswalk, as shown in \cref{fig:pedcar_PC1}. Note that the actual pedestrian position and orientation is shown in red and the vehicle's perceived position and orientation is in gray. In PC2, the expression also includes some prescribed noise that causes the AV to misjudge the location of the pedestrian as shown in \cref{fig:pedcar_PC2}. We conclude from these results that our approach performs well with continuous disturbance spaces and complex probability models. It is able to find STL expressions that reliable produce likelier failures than the importance sampling baseline. 

\begin{table}
    \centering
    \caption{Comparing interpretable validation (IV) to importance sampling (IS) for two different models of the pedestrian crosswalk scenario.}
    \label{tab:pedcar_results}
    \begin{tabular}{@{}llll@{}} 
        \toprule
        \textbf{Method} & \textbf{Fail Rate} & \textbf{Log-Likelihood} & \textbf{Expression}\\
        \midrule
        IV (PC1) & $1.0 \pm 0.0$ & $\num{9.4e-3} \pm 0.0$ & \parbox[t]{2.1cm}{$\square (a_x = 0.019$\\$ \land \ a_y = -0.026)$} \\
        IS (PC1) & $0.13 \pm 0.04$ & $\num{5.2e-32} \pm \num{5.2e-31}$ & N/A \\
        \midrule
        IV (PC2) & $1.0 \pm 0.0$ & $\num{1.3e-23} \pm \num{7.4e-39}$ & \parbox[t]{2cm}{$\square (a_x = -0.32$\\$\land \  a_y = -0.27$ \\ $\land \  n_{v_y} = 0.26)$} \\
        IS (PC2) & $0.08 \pm 0.01$ & $\num{6.8e-56} \pm \num{6.8e-55}$ & N/A \\
        \bottomrule
    \end{tabular}
    \vskip -0.2in
\end{table}

\section{Conclusion}
\label{sec:conclusion}

We presented an approach for the interpretable safety validation of an autonomous system based on signal temporal logic. The approach uses genetic programming to optimize an STL expression so that it describes disturbances trajectories that cause an autonomous system to fail. We demonstrated the approach on two autonomous driving scenarios with discrete and continuous disturbance spaces. We also demonstrated that the technique works in the simple case of disturbances that are independent at each timestep, and in the more complex situation where the disturbances are jointly distributed as a Gaussian process. Compared to an importance sampling baseline, our approach found an order of magnitude more failures of the autonomous vehicle and those failures had a higher likelihood. The failure descriptions helped identify flaws in the autonomoous vehicle under test. Future work will include techniques for ensuring good coverage of the disturbance space, improved optimization techniques, and more complex simulators.

\section*{Acknowledgment}
The authors gratefully acknowledge the financial support from the Stanford Center for AI Safety.

\renewcommand*{\bibfont}{\footnotesize}
\printbibliography
\end{document}